\documentclass{article}

\usepackage[margin=1in]{geometry}
\usepackage{url}            
\usepackage{booktabs}       
\usepackage{amsfonts}       
\usepackage{nicefrac}       
\usepackage{microtype}      
\usepackage{xcolor}         
\usepackage{graphicx}
\usepackage{subfigure}
\usepackage{amsmath}
\usepackage{amssymb}
\usepackage{float}
\usepackage{algorithmic}
\usepackage{algorithm}
\usepackage{bm}
\usepackage[round,semicolon]{natbib}
\usepackage{balance}
\usepackage{multirow}
\usepackage{dcolumn}
\usepackage{textcomp}

\usepackage{multicol}
\newcolumntype{d}{D{.}{.}{-1}}
\usepackage{hyperref}

\definecolor{darkblue}{rgb}{0, 0.2, 0.7}
\hypersetup{
    colorlinks = true,
    linkcolor = darkblue,
    anchorcolor = darkblue,
    citecolor = darkblue,
    filecolor = darkblue,
    urlcolor = darkblue
}

\usepackage{cleveref}

\begin{document}
\twocolumn

\title{\vspace{-2em}%
  \hrule height 4pt%
  \vskip 0.25in%
  \vskip -\parskip%
  \textbf{
  HyperPrompt: Prompt-based Task-Conditioning of Transformers
  }%
  \vskip 0.2in%
  \vskip -\parskip%
  \hrule height 1pt%
  \vskip 0.09in}

\author{
\textbf{Yun He}\thanks{Equal contribution. Yun returned to TAMU, work done as an Intern at Google.}  \quad  
\textbf{Huaixiu Steven Zheng}$^{*}$ \textsuperscript{$\clubsuit$} \quad
\textbf{Yi Tay} \textsuperscript{$\clubsuit$} \quad 
\textbf{Jai Gupta}  \quad \\
\textbf{Yu Du}  \quad
\textbf{Vamsi Aribandi}  \quad
\textbf{Zhe Zhao} \quad
\textbf{YaGuang Li} \quad
\textbf{Zhao Chen} $^{\dagger}$ \quad \\
\textbf{Donald Metzler} \quad
\textbf{Heng-Tze Cheng} \quad
\textbf{Ed H. Chi} \quad \\ \\ Google Research, $^{\dagger}$ Waymo LLC \quad \\ \\
\textsuperscript{$\clubsuit$} \texttt{\{stevenzheng, yitay\}@google.com}
}

\date{}

\maketitle
\sloppy

\begin{abstract}
Prompt-Tuning is a new paradigm for finetuning pre-trained language models in a parameter-efficient way. Here, we explore the use of HyperNetworks to generate hyper-prompts: we propose HyperPrompt, a novel architecture for prompt-based task-conditioning of self-attention in Transformers. The hyper-prompts are end-to-end learnable via generation by a HyperNetwork. HyperPrompt allows the network to learn task-specific feature maps where the hyper-prompts serve as task global memories for the queries to attend to, at the same time enabling flexible information sharing among tasks. We show that HyperPrompt is competitive against strong multi-task learning baselines with as few as $0.14\%$ of additional task-conditioning parameters, achieving great parameter and computational efficiency. Through extensive empirical experiments, we demonstrate that HyperPrompt can achieve superior performances over strong T5 multi-task learning baselines and parameter-efficient adapter variants including Prompt-Tuning and HyperFormer++ on Natural Language Understanding benchmarks of GLUE and SuperGLUE across many model sizes.
\end{abstract}

\begin{figure}[ht]
\centering
\includegraphics[width=8cm]{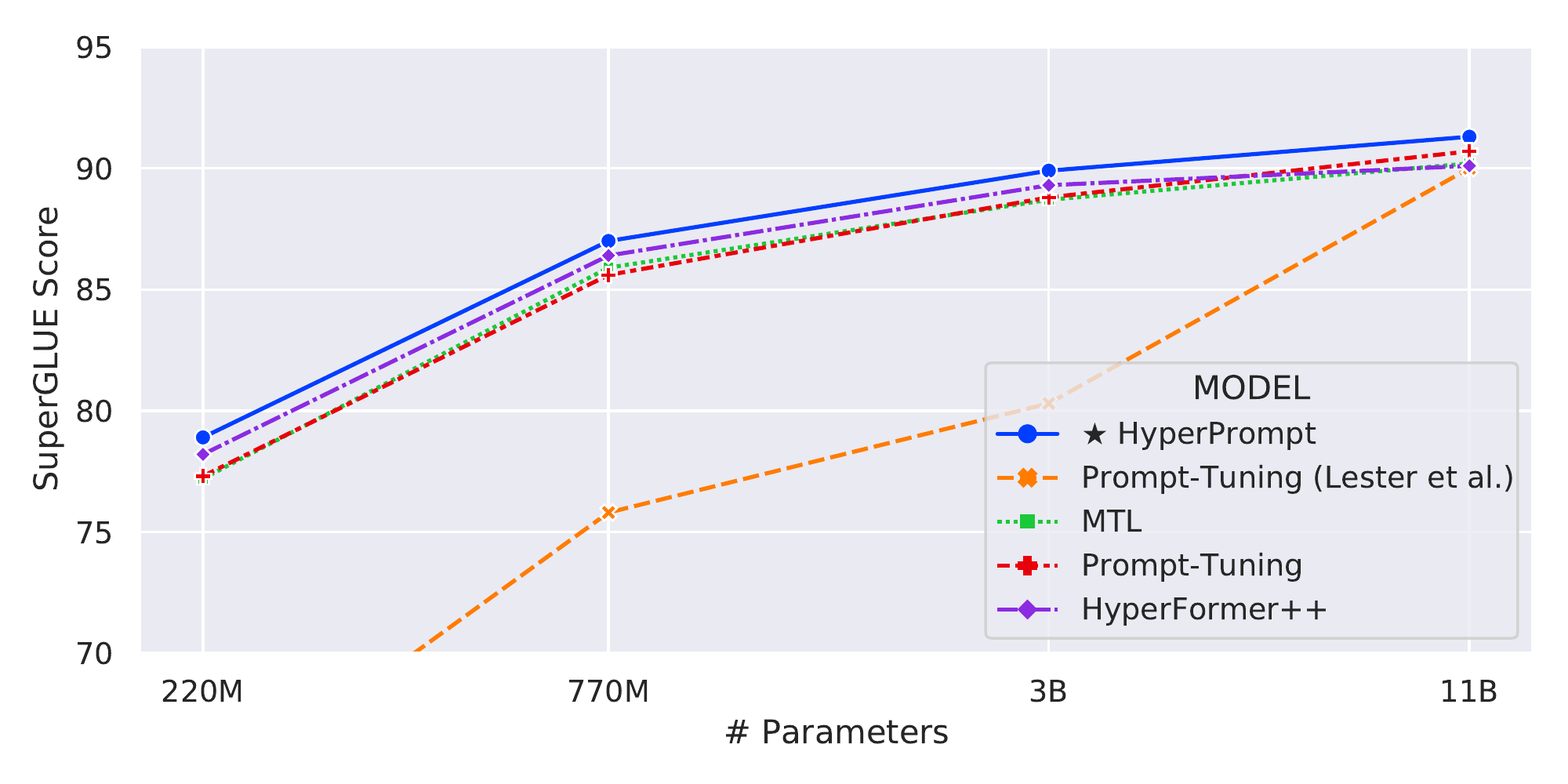}
\caption{HyperPrompt achieves state-of-the-art performance on SuperGLUE for T5 models up to XXL. Prompt-tuning \cite{lester2021power} with tuning prompt parameters only achieves competitive performance against multi-task learning (MTL) baseline for the 11B parameter model with a big performance gap for smaller models. HyperPrompt-Global outperforms the strong parameter-efficient adapter variant HyperFormer++ \cite{karimi-mahabadi-etal-2021-parameter}, the MTL baseline, and the full fine-tuning of Prompt-Tuning (our implementation) across model sizes with a large margin [e.g. 91.3 vs 90.2 (MTL) for T5 XXL].}
\label{fig: hyperprompt_scaling}
\end{figure}

\section{Introduction}
\label{intro}
Prompt-Tuning \citep{lester2021power}, learning to condition large language models with soft learnable memory tokens, have recently garnered attention owing to their ability for parameter-efficient finetuning. Prompts are lightly tuned, allowing the model to be trained quickly since the main body of the pretrained model is kept frozen. To this end, this paradigm is strongly reminiscent of adapter layers \citep{houlsby2019adapter,karimi-mahabadi-etal-2021-parameter,zaken2021bitfit, he2021unified} which are also efficiently finetuned. 

We introduce HyperPrompt, a natural but novel extension of Prompt-Tuning to multi-task learning (MTL) for language. HyperPrompt introduces task-conditioned hyper-prompts that conditions the model on task-specific information for constructing these prompts. Hyper-prompts are injected to the keys and values in the self-attention module, reminiscent of memory augmented Transformers \citep{sukhbaatar2019augmenting}. This mitigates the cost of having prompts pass through the standard FFN layers in Transformers and serves as additional task-specific memory tokens for queries to attend to. 

We further improve upon this by introducing task-aware and layer-aware HyperNetworks \citep{ha2017HyperNetwork} that parameterize and generate weights for the prompt generation process. The usage of HyperNetwork imbues our model with the necessary flexibility and expressiveness, especially when it comes to incorporating task-specific and layer-specific information to the network. Meanwhile, HyperPrompt remains very parameter and computational efficient and friendly to multi-task scaling: the additional parameters scale sub-linearly with, and are independent of the number of tasks in practice. While Hypernetworks have enjoyed some success in learning adapters \citep{karimi-mahabadi-etal-2021-parameter,tay2020hypergrid} and/or continual learning \citep{von2019continual}, we note that this is the first exploration of HyperNetworks as a prompt generator.

 Contrary to prior work, we additionally propose to finetune the entire network instead of only the hyper-prompts. We make several compelling arguments for this. Firstly, \citet{lester2021power} shows that parameter efficient Prompt-Tuning only shines for large (e.g., 11B) models and substantially pales in comparison to fine-tuning when the model is moderately parameterized (e.g., 220M). Secondly, finetuning only adaptive parameters (e.g., prompts/adapters) simply presents an illusion of efficiency \citep{dehghani2021efficiency}. In reality, the FLOPs incurred by the model is still identical on the forward pass, which saves no compute during inference. 
 Parameter counts, especially when including only prompts and adapters, are not the only measurement of computational efficiency. Instead, the FLOPs and training time should be considered together to provide a holistic view.


\paragraph{Our Contributions} Overall, the main contributions include:
\vspace{-1ex}
\begin{itemize}
\vspace{-1ex}
\item We propose a novel HyperPrompt Transformer architecture with learnable hyper-prompts for multi-task fine-tuning with great parameter and computational efficiency.
\vspace{-1ex}
\item We demonstrate that for difficult tasks, it is crucial to fine-tune the task-specific parameters together with the backbone model to achieve Pareto efficiency on all tasks.
\vspace{-1ex}
\item We explore HyperNetworks as a prompt generator, and inject hyper-prompts into the self-attention module as global task memory tokens.
\vspace{-1ex}
\item HyperPrompt outperforms state-of-the-art parameter-efficient T5 models \cite{raffel2019exploring} using Prompt-Tuning or adapters on well-established benchmarks such as SuperGLUE and GLUE, across all explored model sizes (see Figure \ref{fig: hyperprompt_scaling}).
\end{itemize}

\section{Problem Statement}

We consider the general setting of multi-task learning for a set of tasks $\{\mathcal{D_{\tau}}\}_{\tau=1}^{T}$, where $T$ is the total number of tasks and $\{\mathcal{D_{\tau}}\} = \{x_{\tau}^{(n)}, y_{\tau}^{(n)}\}_{n=1}^{N_{\tau}}$ indicates the corresponding training set of the $\tau$-th task with $N_{\tau}$ samples. We assume that a pre-trained Transformer model $f_{\theta}(\cdot)$ (e.g., T5) is given, where the model is parameterized by $\theta$. To tackle such multi-task learning problem with $f_{\theta}(\cdot)$, we minimize the following objective function $\mathcal{L}(\theta) = \sum_{\tau=1}^{T}\sum_{n=1}^{N_{\tau}}C(f_{\theta}(x_{\tau}^{(n)}), y_{\tau}^{(n)})$,
where $C(\cdot , \cdot)$ is typically the cross-entropy loss and $f_{\theta}(x_{\tau}^{(n)})$ is the output for training sample $x_{\tau}^{(n)}$.

Transformer-based pre-trained language models such as T5 \cite{raffel2019exploring} and BART \cite{lewis2020bart} are unified text-to-text frameworks where all tasks share the same encoder-decoder architecture -- $\{\{x_{\tau}^{(n)}\}_{n=1}^{N_{\tau}}\}_{\tau=1}^{T}$ are fed into the same encoder and  $\{\{\hat{y}_{\tau}^{(n)}\}_{n=1}^{N_{\tau}}\}_{\tau=1}^{T}$ are generated by the same decoder. For such universal modules, multi-task learning simply corresponds to mixing task data sets together and there is no task-specific classification or regression networks for each task as in encoder-only modules \citet{devlin2019bert, liu2019multi}. 

Previous work \citet{raffel2019exploring} shows that co-learning all tasks together on a pre-trained Transformer model is inferior to fine-tuning on each task separately. A possible reason is that $\theta$ is task-agnostic (i.e., all parameters are shared) and hence task-specific information is not well captured which can be especially true for low-resource tasks. Therefore, a natural way to improve the performance of Transformers on multi-task learning is to introduce a set of task-conditioned parameters $\{\delta_{\tau}\}_{\tau=1}^{T}$ into $f_{\theta}(.)$. The objective function can be updated as $\mathcal{L}(\theta, \{\delta_{\tau}\}_{\tau=1}^{T}) = \sum_{\tau=1}^{T}\sum_{n=1}^{N_{\tau}}C(f_{\theta, \delta_{\tau}}(x_{\tau}^{(n)}), y_{\tau}^{(n)})$,
where $\delta_{\tau}$ is the task-specific parameterization for the $\tau$-th task. During training, both $\theta$ and $\{\delta_{\tau}\}_{\tau=1}^{T}$ are updated via back-propagation because we observe a large performance drop in SuperGLUE when backbone model $\theta$ is frozen and only  task-conditioned parameters are tuned, as done in \citet{karimi-mahabadi-etal-2021-parameter}, which will be detailed in Section \ref{exp: Tuning all vs Task-Conditioned Parameters only}.

To this end, our goal is to design task-conditioned parameterization of Transformer models to achieve \textit{greater parameter and computational efficiency as well as Pareto efficiency for multi-task learning}. More explicitly, we have two goals: (1) improving the finetuning performance of most tasks in $\{\mathcal{D_{\tau}}\}_{\tau=1}^{T}$ by introducing task-conditioned parameters $\{\delta_{\tau}\}_{\tau=1}^{T}$ into $f_{\theta}(.)$ and (2) under the constraint that $\sum_{\tau}\| \{\delta_{\tau}\}_{\tau=1}^{T} \|_{0} \ll \| \theta \|_{0}$, which means that the model capacity will not be significantly increased. And the computational cost would not increase substantially either. 

\section{Methods}
\begin{figure*}[ht]
\centering
\includegraphics[width=15cm]{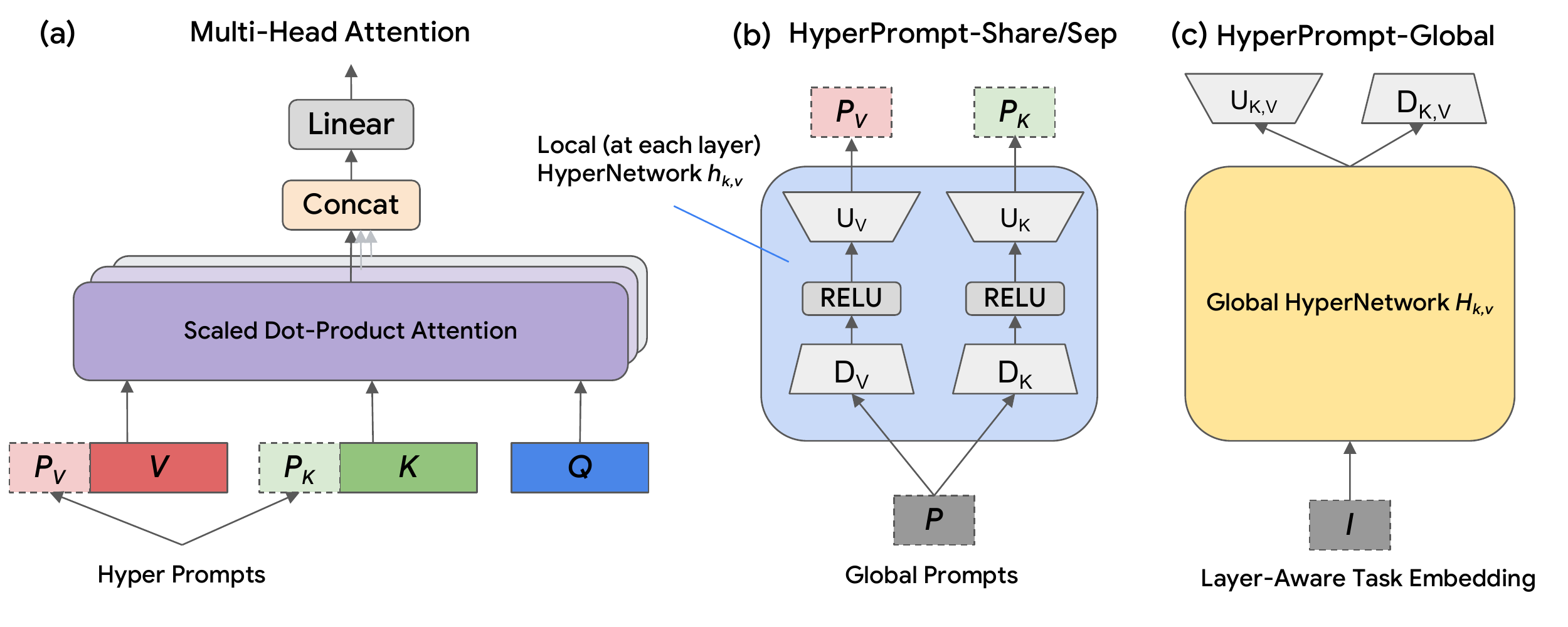}
\caption{\hangindent=0.7cm HyperPrompt framework: (a) in each Transformer block, task-specific hyper-prompts $P_{K,V}$ are prepended to the original key $K$ and value $V$ for the query $Q$ to attend to, (b) in HyperPrompt-Share/Sep, global prompts $P$ are used to generate the hyper-prompts $P_{K,V}$ through local HyperNetworks $h_{k,v}$ at each Transformer layer, which consists of a down-projection matrix $D_{K,V}$, a RELU layer and a up-project matrix $U_{K,V}$, (c) in HyperPrompt-Global, all the local HyperNetworks ($D_{K,V}$, $U_{K,V}$) are generated by global HyperNetworks $H_{k,v}$ using layer-aware task embeddings $I$ as task-specific inputs (see Section \ref{sec: HyperPrompt} for details).}
\label{fig:schematic_hyperprompt}
\end{figure*}

In this section, we introduce HyperPrompt which has three variants: HyperPrompt-Share, HyperPrompt-Sep and HyperPrompt-Global (Figure \ref{fig:schematic_hyperprompt}). We follow two key design principles to formulate HyperPrompt: (1) injecting task-conditioning into self-attention module for better computational efficiency and more expressive power via token-level interactions, and (2) using HyperNetworks to simultaneously improve the parameter efficiency and allow a flexible degree of task sharing for better generalization. 

\subsection{Prompt-Based Task-Conditioned Transformer}
Previous adapter-based methods \cite{karimi-mahabadi-etal-2021-parameter, tay2020hypergrid} for multi-task learning normally add an adapter (i.e., dense-relu-dense network) for each task after the feed-forward layers at every Transformer block. Instead, the key idea of our approach is to prepend $l$ task-conditioned trainable vectors to the keys and values of the multihead self-attention layer at every Transformer block, where the task-specific attention feature maps are jointly learned with the task-agnostic representation. 

The idea of prepending learnable prompts to the network is explored before by \citet{li2021prefix, lester2021power, liu2021gpt} for single-task fine-tuning. We first introduce and expand this idea for multi-task learning in this subsection. Specifically, we design a novel method called HyperPrompt following the design principle $\#1$ of injecting hyper-prompts into self-attention and $\#2$ using HyperNetworks as generators for hyper-prompts.

At a multihead self-attention layer, the original key, value and query are calculated as $\bm{K}_{\tau}=\bm{X}_{\tau}\bm{W}_{k}$, $\bm{V}_{\tau}=\bm{X}_{\tau}\bm{W}_{v}$, $\bm{Q}_{\tau}=\bm{X}_{\tau}\bm{W}_{q}$,
where $\bm{X}_{\tau} \in \mathbb{R}^{L \times d}$ is the input sequence of a training sample from the $\tau$-th task, $L$ is the sequence length, $d$ is the model dimension. $\bm{W}_{k} \in \mathbb{R}^{d \times h \times d_{h}}$, $\bm{W}_{v} \in \mathbb{R}^{d \times h \times d_{h}}$ and $\bm{W}_{q} \in \mathbb{R}^{d \times h \times d_{h}}$ project the input into original key $\bm{K}_{\tau} \in \mathbb{R}^{L \times h \times d_{h}}$, value $\bm{V}_{\tau} \in \mathbb{R}^{L \times h \times d_{h}}$ and query $\bm{Q}_{\tau} \in \mathbb{R}^{L \times h \times d_{h}}$, $h$ is the number of heads, $d_{h}$ is the dimension of each head and typically set to $d/h$ to save parameters.

To learn the task-specific information for the $\tau$-th task, we have $l$ trainable $d$-dimensional vectors as the hyper-prompts for the key and the value respectively, denoted as $\bm{P}_{\tau, k} \in \mathbb{R}^{l \times h \times d_{h}}$ and $\bm{P}_{\tau, v} \in \mathbb{R}^{l \times h \times d_{h}}$, as shown in Figure \ref{fig:schematic_hyperprompt}(a). Then, the hyper-prompts are concatenated with the original key and value:
\begin{equation}
\label{equ: concat with key}
    \bm{K^{\prime}}_{\tau} = \text{concat}(\bm{P}_{\tau, k}, \: \bm{K}_{\tau})
\end{equation}
\begin{equation}
\label{equ: concat with value}
    \bm{V^{\prime}}_{\tau} = \text{concat}(\bm{P}_{\tau, v}, \: \bm{V}_{\tau})
\end{equation}
where the new key (value) $\bm{K^{\prime}}_{\tau}$ ($\bm{V^{\prime}}_{\tau}) \in \mathbb{R}^{(l+L) \times h \times d_{h}}$ are used to compute the multihead self-attention.

After that, the multihead self-attention can be operated:
$\bm{O}_{\tau} = \text{Attention}(\bm{Q}_{\tau}, \bm{K^{\prime}}_{\tau}, \bm{V^{\prime}}_{\tau}) = \text{softmax}(\bm{Q}_{\tau}\bm{K^{\prime T}}_{\tau})\bm{V^{\prime}}_{\tau}$
where $\bm{O}_{\tau} \in \mathbb{R}^{L \times d}$ is the output of multihead attention. 

The hyper-prompts benefit Transformers for multi-task learning in two ways:
(1) Prompt for key $\bm{P}_{\tau, k}$ is prepended with the original key and will participate in the calculation of attention feature map: $\text{softmax}(\bm{Q}_{\tau}\bm{K^{\prime T}}_{\tau})$. $\bm{P}_{\tau, k}$ directly interacts (matrix multiplication) with the original query $\bm{Q}_{\tau}$, allowing tokens to acquire task-specific semantics.
(2) 
Prompt for value $\bm{P}_{\tau, v}$ is prepended with the original value and will be absorbed into the self-attention output $\bm{O}_{\tau}$, where each position in $\bm{O}_{\tau}$ is the weighted-sum of vectors in $\bm{V^{\prime}}_{\tau}$ with weights from the attention scores. This way, $\bm{P}_{\tau, v}$ can serve as task-specific memories for multihead attention to retrieve information from.

\subsection{HyperPrompt}
How to obtain the prompts for the $m$-th Transformer block? A straightforward way is to directly initialize $\bm{P}^{m}_{\tau, k}$ and $\bm{P}^{m}_{\tau, v}$. However, this way is parameter-inefficient, as it scales linearly with both the number of tasks $T$ and the number layers $M$ as $\mathcal{O}(T\times M)$.

Instead, we initialize a global\footnote{we term it \textit{global} because it is independent of the layer number as opposed to layer-dependent prompt $\bm{P}^{m}_{\tau}$.} prompt $\bm{P}_{\tau}$ for each task and apply local HyperNetworks at every Transformer block to project this prompt into $\{ \bm{P}^{m}_{\tau, k} \}_{m=1}^{M}$ and $\{ \bm{P}^{m}_{\tau, v} \}_{m=1}^{M}$.

\textbf{Global Prompts.} Specifically, we initialize a set of global prompts $\{\bm{P}_{\tau}\}_{\tau=1}^{T}$, where $\bm{P}_{\tau} \in \mathbb{R}^{l \times d}$ is a trainable matrix to learn the task-specific information of the $\tau$-th task, $d$ is the model dimension and $l$ is the length of the prompt.

\textbf{Local HyperNetworks.} At the $m$-th Transformer block, we apply two local HyperNetworks $h_{k}^{m}$ and $h_{v}^{m}$ to transform the global prompt $\bm{P}_{\tau}$ into layer-specific and task-specific prompts as shown in Figure \ref{fig:schematic_hyperprompt}(b):
\begin{equation}
\label{equ: shared key projection network}
    \bm{P}^{m}_{\tau, k} = h_{k}^{m}(\bm{P}_{\tau}) = \bm{U}_{k}^{m} (\text{Relu}(\bm{D}_{k}^{m}(\bm{P}_{\tau}))),
\end{equation}
\begin{equation}
\label{equ: shared value projection network}
    \bm{P}^{m}_{\tau, v} = h_{v}^{m}(\bm{P}_{\tau}) = \bm{U}_{v}^{m} (\text{Relu}(\bm{D}_{v}^{m}(\bm{P}_{\tau}))),
\end{equation}
where $\bm{P}^{m}_{\tau, k/v} \in \mathbb{R}^{l \times h \times d_{h}}$. We call these generated prompts \textit{hyper-prompts} to distinguish from global prompts.

In particular, to limit the number of parameters, the local HyperNetworks are designed using a bottleneck architecture:  $\bm{D}_{k/v}^{m} \in \mathbb{R}^{d \times b}$ and $\bm{U}_{k/v}^{m} \in \mathbb{R}^{b \times h \times d_{h}}$ are down-projection and up-projection matrices, respectively. $b$ is the bottleneck dimension satisfying $b \ll d$.

\textbf{HyperPrompt-Share.} We first have all tasks share the same two local HyperNetworks defined by the down-project matrices $\bm{D}_{k}^{m}$ and $\bm{D}_{v}^{m}$, and the up-project matrices $\bm{U}_{k}^{m}$ and $\bm{U}_{v}^{m}$. We refer to this design choice as HyperPrompt-Share.

Despite the saving of parameters, one drawback of HyperPrompt-Share is that the task conflicts could arise given the limited model capacity \cite{Wu2020Understanding, wang2020small} of the shared local HyperNetworks.

\textbf{HyperPrompt-Sep.} In the opposite extreme of HyperPrompt-Share, each task can have its own local HyperNetworks $h_{\tau, k}^{m}(\bm{P}_{\tau})$ and $h_{\tau, v}^{m}(\bm{P}_{\tau})$ as following:
\begin{equation}
\label{equ: sep key projection network}
    \bm{P}^{m}_{\tau, k} = h_{\tau, k}^{m}(\bm{P}_{\tau}) = \bm{U}_{\tau, k}^{m} (\text{Relu}(\bm{D}_{\tau, k}^{m}(\bm{P}_{\tau}))),
\end{equation}
\begin{equation}
\label{equ: sep value projection network}
    \bm{P}^{m}_{\tau, v}  = h_{\tau, v}^{m}(\bm{P}_{\tau}) = \bm{U}_{\tau, v}^{m} (\text{Relu}(\bm{D}_{\tau, v}^{m}(\bm{P}_{\tau}))),
\end{equation}
where $\bm{D}_{\tau, k/v}^{m}$ and  $\bm{U}_{\tau, k/v}^{m}$ are down-projection and up-projection matrices for the $\tau$ task, respectively.
In this case, each task hyper-prompt is trained independently and hence there is no information sharing.

\subsection{HyperPrompt-Global}
\label{sec: HyperPrompt}
We further propose a novel design of \textit{HyperPrompt-Global} to flexibly share information and knowledge among tasks and blocks while maintaining a low parameter cost. As shown in Figure \ref{fig:schematic_hyperprompt}(c), the key idea of HyperPrompt-Global is to generate the local HyperNetworks using the same global HyperNetwork shared by all tasks and all Transformer blocks.



\textbf{Layer-Aware Task Embedding.} Following the same recipe in \citet{karimi-mahabadi-etal-2021-parameter}, we define a layer-aware task embedding for better generalization. Let $k_{\tau} \in \mathbb{R}^{t^{\prime}}$ denote the task embedding for the $\tau$ task and $t^{\prime}$ is the dimension.
To capture the layer-specific information, layer embedding $z_{m} \in \mathbb{R}^{t^{\prime}}$ is introduced.
After that, a task projection network $h_{t}(\cdot, \cdot)$ is applied to fuse the task embedding and the layer embedding into the final layer-awared task embedding $\bm{I}_{\tau}^{m} = h_{t}(k_{\tau}, z_{m})$,
where $\bm{I}_{\tau}^{m}$ is the input to the shared global HyperNetworks as shown in Figure \ref{fig: hyperprompt_scaling}(c). $h_{t}$ is a MLP consisting of two feed-forward layers and a ReLU non-linearity, which takes the concatenation of $k_{\tau}$ and $z_{m}$ as input.

\textbf{Global HyperNetworks.}  $H_{k}(\cdot)$ generates the weight matrices $(\bm{U}_{\tau, k}^{m}, \bm{D}_{\tau, k}^{m})$ in the local HyperNetworks of key hyper-prompts and another global HyperNetwork $H_{v}(\cdot)$ generates the weight matrices $(\bm{U}_{\tau, v}^{m}, \bm{D}_{\tau, v}^{m})$ in the local HyperNetworks of value hyper-prompts:

\begin{equation}
\label{equ: global HyperNetwork k}
    (\bm{U}_{\tau, k}^{m}, \bm{D}_{\tau, k}^{m}) = H_{k}(\bm{I}_{\tau}^{m}) = (\bm{W}^{U_{k}}, \bm{W}^{D_{k}})\bm{I}_{\tau}^{m},
\end{equation}
\begin{equation}
\label{equ: global HyperNetwork v}
    (\bm{U}_{\tau, v}^{m}, \bm{D}_{\tau, v}^{m}) = H_{v}(\bm{I}_{\tau}^{m}) = (\bm{W}^{U_{v}}, \bm{W}^{D_{v}})\bm{I}_{\tau}^{m},
\end{equation}

where $\bm{I}_{\tau}^{m} \in \mathbb{R}^{t}$ is the layer-aware task embedding for the $\tau$ task at the $m$-th block.
$\bm{W}^{D_{k}} \in \mathbb{R}^{(d \times b) \times t}$, $\bm{W}^{D_{v}} \in \mathbb{R}^{(d \times b) \times t}$, $\bm{W}^{U_{k}} \in \mathbb{R}^{(b \times h \times d_{h}) \times t}$ and $\bm{W}^{U_{v}} \in \mathbb{R}^{(b \times h \times d_{h}) \times t}$ are the weight matrices of $H_{k}(\cdot)$ and $H_{v}(\cdot)$. 

Given that $\bm{U}_{\tau, k/v}^{m}$, and $\bm{D}_{\tau, k/v}^{m}$ are generated by the global HyperNetworks, we project the global prompts $\bm{P}_{\tau, k/v}$ into hyper-promtps $\bm{P}^{m}_{\tau, k/v}$ following Eqs. \ref{equ: sep key projection network} and \ref{equ: sep value projection network}. Finally, the hyper-prompts $\bm{P}^{m}_{\tau, k/v}$ are prepended with original key and value at every self-attention layer as shown in Figure \ref{fig:schematic_hyperprompt}(a)  to calculate the task-conditioned attention scores.

Using global HyperNetworks to generate the projection networks has two benefits:
\begin{enumerate}
    \item It enables a more flexible way to share information across tasks and layers: the transformation matrices are decomposed into $H_{k/v}(\cdot)$ that are shared by all tasks and all layers.
    Therefore, the model can adjust the degree of information sharing across tasks and layers through learning the appropriate parameter values in $H_{k/v}(\cdot)$ during the end-to-end training. 
    \item A parameter-efficient task conditioned parameterization is enabled. The number of extra task-conditioned parameters doesn't depend on the number of layers $M$, and scales sub-linearly with respect to the total number of tasks $T$. In practice, since task embeddings and task prompts have far fewer parameters than the global HyperNetworks, the additional task-conditioned parameters is almost independent of $T$. 
\end{enumerate}


\subsection{Parameter Efficiency of HyperPrompt}
\label{subsec: para-efficiency}

As shown in \ref{appendix: para-efficiency}, the total number of additional parameters from HyperPrompt-Global is $dlT+4(bdt)+Tt^{\prime}+Mt^{\prime}+(2t^{\prime} + t)e$, where $d$ is the model dimension, $l$ is the length of the prompts, $T$ is the total number of tasks, $b$ is the bottleneck dimension of the weight matrices of the local HyperNetworks, $d$ is the model dimension, $t^{\prime}/t$ is the dimension of the raw/final layer-aware task embedding, and $e$ is the hidden dimension of $h_{k/v}$. Therefore, the space complexity is $\mathcal{O}(d(lT+4bt))$, given that in practice $M\sim T$, $t^{\prime}\ll dl$, and $e\ll bd$. This leads to a sub-linear scaling with respect to $T$.

Furthermore, $T$ is typical $\sim \mathcal{O}(10)$ for multi-task learning. A reasonable $l\sim \mathcal{O}(10)$ is required to achieve the optimal performance, which will be detailed in Section \ref{sec: Impact of Prompt Length}. On the other hand, typical values for $b\sim 24$ and $t\ge 32$, and therefore $4bt \gg lT$ is satisfied in most cases. Hence, the space complexity could be further simplified as $\mathcal{O}(bdt)$. In conclusion, the space complexity of HyperPrompt-Global mainly comes from the global HyperNetworks and is practically independent of the prompt length $l$, the number of Transformer layers $M$, and the number of tasks $T$.

\section{Experiments}

\subsection{Experimental Setup}
\label{sec: Experimental Setup}
\textbf{Datasets.} We evaluate the performance of the models on GLUE \cite{wang2018glue} and SuperGLUE \cite{wang2019superglue} respectively. Each of them is a collection of text classification tasks to test the general language understanding ability. Specifically, the tasks include: sentence acceptability (CoLA), sentiment analysis (SST-2), paraphrasing/sentence similarity (MRPC, STS-B and QQP), natural language inference (MNLI, QNLI, RTE and CB), coreference resolution (WSC), sentence completion (COPA), word sense disambiguation (WIC) and question answering (MultiRC and ReCoRD, BoolQ).

\textbf{Transformers.} Following previous work \citet{karimi-mahabadi-etal-2021-parameter} and \citet{tay2020hypergrid}, our models are built on top of the state-of-the-art Transformer model T5 \cite{raffel2019exploring}, which uses encoder-decoder architecture from \citet{vaswani2017attention}. We use already pre-trained T5 with sizes from Base (220M parameters) to XXL (11B). 

\textbf{Evaluation.} We save a checkpoint every 2000 steps for all models and  follow the same convention as \citet{raffel2019exploring} in selecting the best checkpoint for each task. The emphasis of our evaluation is not to find the best single checkpoint for all tasks but to test the model's ability of transfer learning among the co-trained tasks. We first calculate the average of all metrics for each task and then report the average of all tasks for GLUE and SuperGLUE.

\textbf{Baselines.} We compare our proposed MTL-Prompt and HyperPrompt-Share/Sep/Global with vanilla T5 models \cite{raffel2019exploring} for multi-task learning, which is referred to \textit{MTL}. Another baseline is \textit{Vanilla Adapter} proposed in \citet{houlsby2019parameter} that add adapters modules for each task after each of the the two feed-forward modules in each Transformer block of the T5 model. The state-of-the-art adapter-based method for multi-task learning is \textit{HyperFormer++} proposed in \citet{karimi-mahabadi-etal-2021-parameter} that use HyperNetworks to generate adapters for each task and add them after the feed-forward modules following \citet{houlsby2019parameter}. In addition, \textit{Prompt-Tuning} \cite{lester2021power} is originally for parameter-efficient single-task fine-tuning and only prepends prompts to the input word embeddings in the first layer. We slightly modify it by initializing and prepending prompts for each task respectively so that \textit{Prompt-Tuning} can be applied to multi-task learning.

We defer additional details of the experiments to \ref{appendix: exp-details}

\subsection{Key Results}
\label{exp: summary}
Figure \ref{fig: hyperprompt_scaling} provides an overall summary of the results of HyperPrompt. Previous prompt-tuning \cite{lester2021power, li2021prefix} methods focus on parameter-efficient single-task fine-tuning and hence freeze the backbone and only fine-tune the prompts. Their experiments show that the performance of only tuning the prompts can match the full model training with a very large 11B model (Figure \ref{fig: hyperprompt_scaling}), but substantially pales for moderate model sizes. 

Our HyperPrompt-Global architecture when fully fine-tuned achieves state-of-the-art performance on SuperGLUE across four different model sizes. Competitive adapter-tuning variants including Prompt-Tuning and HyperFormer++ can either match or slightly improve upon the multi-task learning (MTL) baseline on the SuperGLUE dataset. In contrast, HyperPrompt-Global outperforms the strong MTL baseline by a large margin on SuperGLUE score ($78.9$ vs $77.2$ for T5 Base). Interestingly, such a performance gain continues all the way to model size as big as XXL (e.g. $91.3$ vs $90.2$) with only $0.14\%$ additional parameters.

\subsection{Tuning all vs Task-Conditioned Parameters}
\label{exp: Tuning all vs Task-Conditioned Parameters only}

\begin{table}
    \centering
    \resizebox{0.48\textwidth}{!}
    {%
    \begin{tabular}{lccc}
    \toprule
    Tunable & Model & GLUE & SuperGLUE \\
    \midrule
    All & MTL & 88.3 & 85.9\\
    All & HyperFormer++ & 88.8 & 86.4\\
    All & HyperPrompt-Global & 89.4 & 87\\
   Task & HyperFormer++ & 87.3 & 80.5 \\
   Task & HyperPrompt-Global & 87.5 & 81.5 \\
    \midrule
        
        \bottomrule
    \end{tabular}
    }%
    \caption{Comparison of fine-tuning all vs task-specific parameters using T5 Large. The average scores of GLUE and SuperGLUE are reported on T5 Large.}
    \label{tab:tune_all_vs_specific}
\end{table}

Recently, \citet{karimi-mahabadi-etal-2021-parameter} show that only tuning adapters can be competitive against the full fine-tuning. However, the evaluation is conducted only on the GLUE with smaller models including T5 Small and Base.

In the experiments, we first compare tuning the full model vs. only task-conditioned parameters. Table \ref{tab:tune_all_vs_specific} shows the comparison on the GLUE and SuperGLUE average scores using T5 large (for per-task performance, please refer to \ref{appendix: per-task-perf}). 
For GLUE, the observation is consistent with \cite{karimi-mahabadi-etal-2021-parameter}, where task-specific only fine-tuning of HyperFormer++ and HyperPrompt-Global is comparable to the MTL baseline. However, on SuperGLUE, we observe a large gap: the average score drops by 5.5 and 5.9 for HyperPrompt-Global and HyperFormer++, respectively.

Therefore, these experiments show that only tuning the task-conditioned parameters is not enough to achieve competitive results as full model training for multi-task learning on high-difficulty tasks such as SuperGLUE. This is consistent with the results of Prompt-Tuning \cite{lester2021power}. Hence, the rest of the experiments are conducted with tuning all model parameters. 

\subsection{Computational Efficiency}
Table \ref{tab:flops_training_time} presents the computational efficiency of the Adapter/Prompt models. HyperPrompt-Global (together with HyperPrompt-Share) has the lowest \# Ops since hyper-prompts are injected into self-attention and skip the standard FFN layers. In contrast, HyperFormer++ has $\sim 3\text{x}$ \# Ops compared to other variants. Regarding training time, HyperPrompt-Share is fastest given that the local HyperNetworks are shared across tasks. Vanilla Adapter and HyperPrompt-Global are comparable while HyperFormer++ and Prompt-Tuning take significant longer to do the full fine-tuning. This shows the computational efficiency of HyperPrompt for both training and inference.

\begin{table}[]
    \centering
    \begin{tabular}{l|c c}
    \toprule
        Model &  \# Ops & Training Time\\
        \midrule
        Vanilla Adapter & $1.01 \times 10^{13}$ & 8.4h \\
        HyperFormer++ & $3.14 \times 10^{13}$ & 10.3h \\
        Prompt-Tuning & $1.16 \times 10^{13}$ & 11.1h\\
        HyperPrompt-Sep & $1.01 \times 10^{13}$ & 8.9h \\
        HyperPrompt-Share & \bm{$9.8 \times 10^{12}$} & \textbf{8.0h} \\
        HyperPrompt-Global & \bm{$9.8 \times 10^{12}$} & 8.7h\\
        \bottomrule
    \end{tabular}
    \caption{The number of operations for a single forward pass and training time on T5 Base.}
    \label{tab:flops_training_time}
\end{table}

\subsection{Ablation Study}
Table \ref{tab:hyperprompt_base} presents the results on T5 Base and Table \ref{tab:hyperprompt_large} presents the results on T5 Large (see more detailed results in \ref{appendix: per-task-perf}). HyperPrompt-Global outperforms all baselines in terms of the average score of GLUE and SuperGLUE.

\textbf{HyperPrompt-Global vs. Prompt-Tuning.} The original Prompt-Tuning \cite{lester2021power} is for single-task fine-tuning. To be parameter-efficient, it only trains the prompts with the backbone frozen. 
To make a fair comparison, we modify Prompt-Tuning by (1) training both prompts and backbone, and (2) adding prompt to each task and co-train all tasks together. As shown in Table \ref{tab:hyperprompt_base} and \ref{tab:hyperprompt_large}, HyperPrompt-Global outperforms Prompt-Tuning by 2.0 (0.6) and 1.6 (1.4) on GLUE and SuperGLUE using T5 Base (Large), respectively. HyperPrompt-Global improves upon Prompt-Tuning in two places: (1) 
Prompt-Tuning only adds prompts to the word embedding layer while HyperPrompt-Global adds hyper-prompts at every Transformer layer and hence is more expressive; and (2) Prompts of tasks are trained independently in Prompt-Tuning while HyperPrompt-Global enables a flexible information sharing via HyperNetworks. 

\begin{table}
    \centering
    \resizebox{0.48\textwidth}{!}
    {%
    \begin{tabular}{l|ccc}
    \toprule
    Model & \#Params & GLUE & SuperGLUE \\
    \midrule
       MTL  & 1.0x & 85.5 (0.9) & 77.2 (0.2)\\
       Vanilla Adapter & 1.06x & 86.7 (0.3) & 77.5 (0.1)\\
       HyperFormer++ & 1.04x & 86.5 (0.0) & 78.2 (0.7)\\
       Prompt-Tuning & 1.0003x & 84.8 (0.6) & 77.3 (0.2)\\
       \midrule
      HyperPrompt-Share & 1.008x & 86.4 (0.6) & 78.2 (0.7)\\
       HyperPrompt-Sep & 1.06x & \textbf{86.8 (0.1)} & 77.5 (0.1)\\
       HyperPrompt-Global & 1.04x & \textbf{86.8 (0.4)} & \textbf{78.9 (0.5)}\\
        
        \bottomrule
    \end{tabular}
    }%
    \caption{GLUE and SuperGLUE average scores (standard deviations) over 3 runs of HyperPrompt against baselines on T5 Base.}
    \label{tab:hyperprompt_base}
\end{table}

\begin{table}[b]
    \centering
    \resizebox{0.48\textwidth}{!}
    {%
    \begin{tabular}{l|ccc}
    \toprule
    Model & \#Params & GLUE & SuperGLUE \\
    \midrule
       MTL  & 1.0x & 88.3 (0.6) & 85.9 (0.3) \\
       Vanilla Adapter & 1.06x & 88.8 (0.2) & 86.1 (0.5) \\
       HyperFormer++ & 1.02x & 88.8 (0.0) & 86.4 (0.5)\\
       Prompt-Tuning & 1.0001x & 88.8 (0.3) & 85.6 (0.1)\\
    \midrule
      HyperPrompt-Share & 1.008x & 89.3 (0.1) & 86.8 (0.2)\\
       HyperPrompt-Sep & 1.06x & \textbf{89.4 (0.2)} & 86.1 (0.3)\\
       HyperPrompt-Global & 1.02x & \textbf{89.4 (0.1)} & \textbf{87.0 (0.5)}\\
        
        \bottomrule
    \end{tabular}
    }%
    \caption{GLUE and SuperGLUE average scores (standard deviations) over 3 runs of HyperPrompt against baselines on T5 Large.}
    \label{tab:hyperprompt_large}
\end{table}

\textbf{HyperPrompt-Global vs. HyperFormer++.} Our method is superior to the state-of-the-art baseline HyperFormer++ in the average score of GLUE and SuperGLUE for both Base and Large T5 model. For example, HyperPrompt-Global of T5 large achieves 87.0 on the SuperGLUE compared to 86.4 by HyperFormer++ (Table \ref{tab:hyperprompt_large}). Note that the main difference between the two methods is that HyperPrompt-Global inserts the task-conditioned parameters as prompts into self-attention layers while HyperFormer++ insert adapters after each block.
We believe task-conditioning in self-attention gives more expressive power than in the feed-forward network as done in adapters. Hyper-prompts that are prepended with the key and value participate in the attention interactions between different token positions, which helps the model to better capture the task-dependent semantics.

\textbf{HyperPrompt-Global vs. MTL.} Next, we observe that using HyperPrompt-Global can greatly improve the performance upon the vanilla Transformer model (referred to MTL): 1.7 (1.1) gain on SuperGLUE score for T5 Base (Large) with $4\%$ ($2\%$) additional paramters. In conclusion, the experiments show that HyperPrompt-Global is a parameter-efficient and effective task-conditioned parameterization of Transformers for multi-task learning.


\textbf{HyperPrompt-Global vs. HyperPrompt-Share/Sep.} Interestingly, HyperPrompt-Share is better than HyperPrompt-Sep on the SuperGLUE on both Base and Large models while the opposite is true for GLUE. Notice that all tasks share the same two projection networks in HyperPrompt-Share while each task has its own projection networks in HyperPrompt-Sep. 
More importantly, we observe that HyperPrompt-Global, where the projection networks are generated by the global HyperNetworks, always achieves the best performance on both GLUE and SuperGLUE. Hence, the experiments show that HyperPrompt-Global can adjust the degree of information sharing for better multi-task generalization, compared to HyperPrompt-Share/Sep.

\subsection{Peeking into Hyper-Prompts}
\label{sec:peeking}
\begin{figure}[t]
  \centering
 \subfigure{
    \includegraphics[width=3.2in]{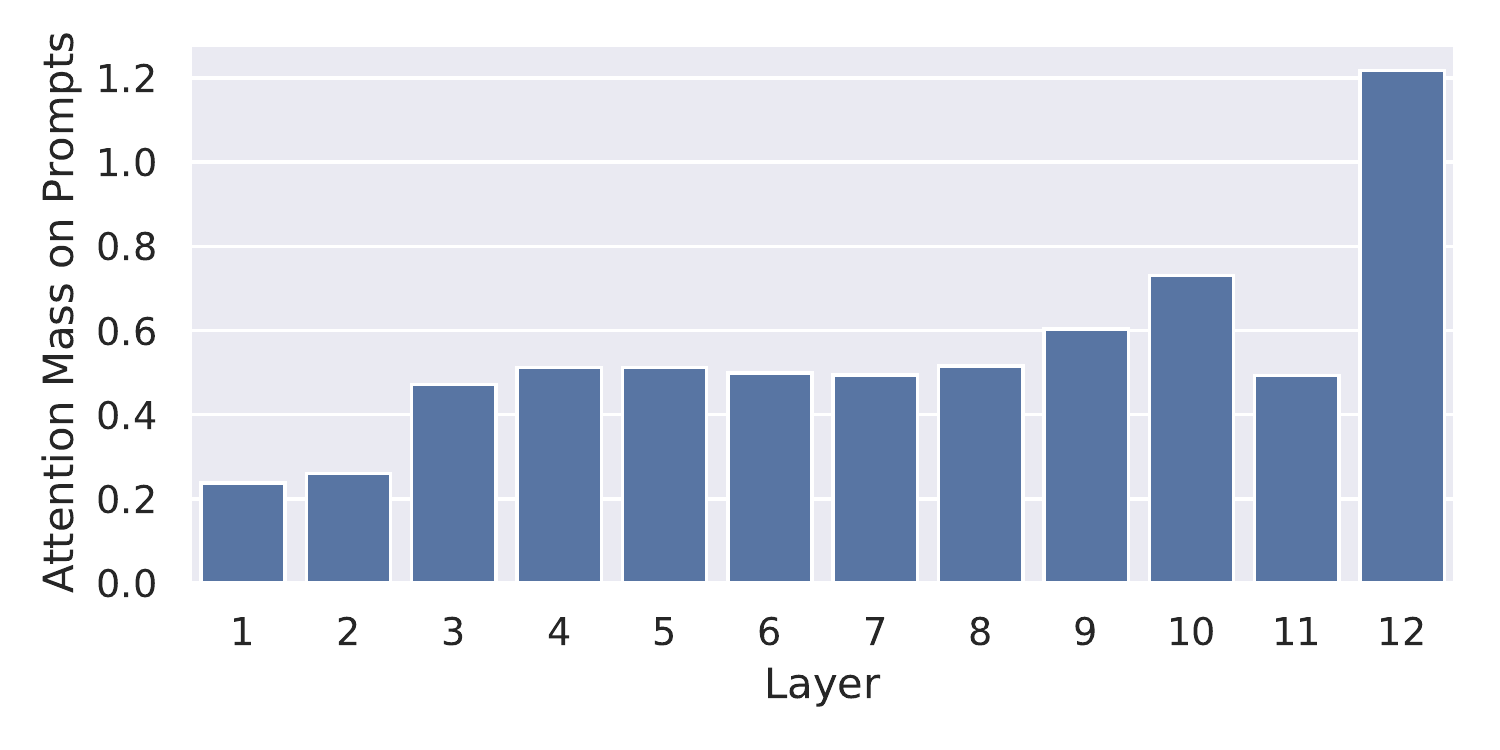}}
  \subfigure{
    \includegraphics[width=3.2in]{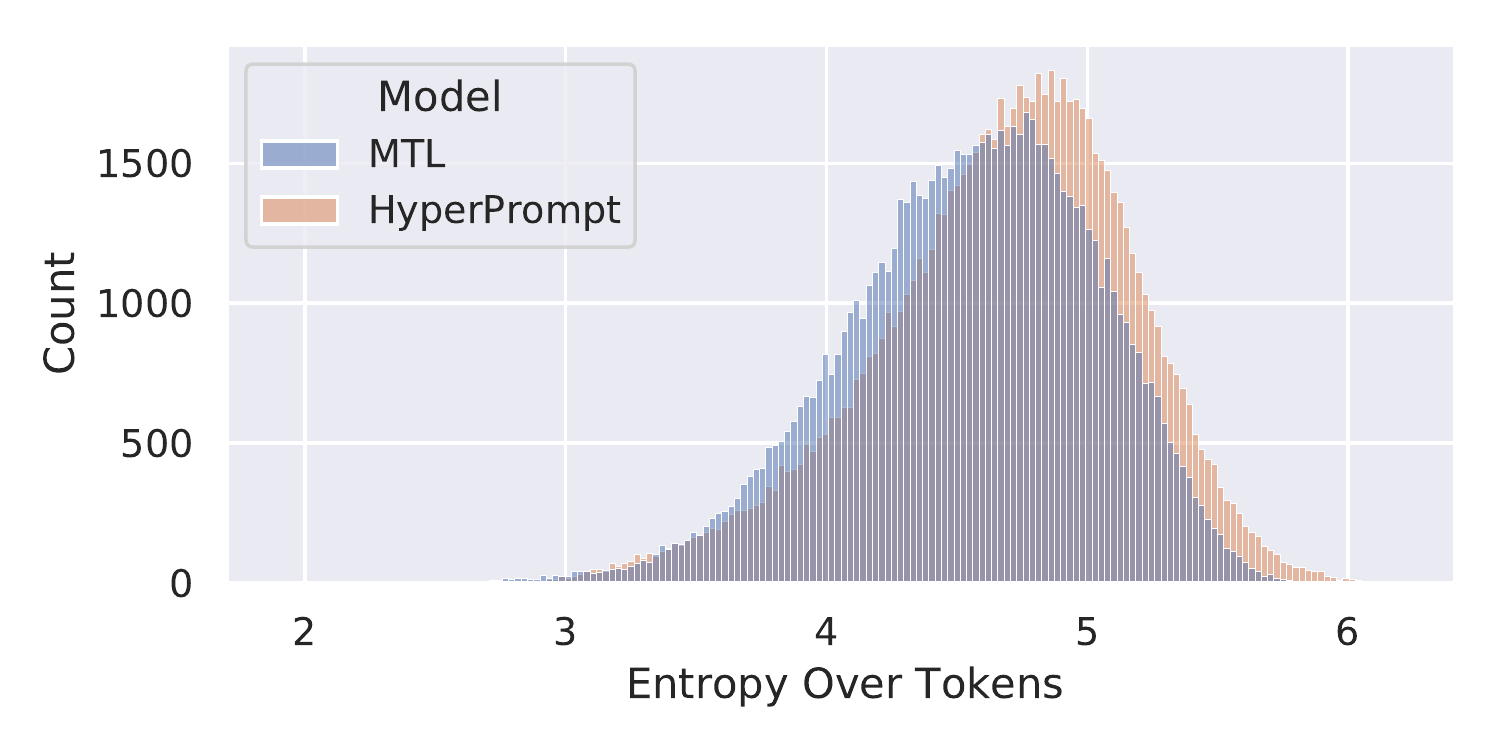}}
   \caption{Visualization of attention mass and entropy distribution.}
  \label{fig: attention_vis} 
\end{figure}

To shed light on how hyper-prompts help improve the multi-task generalization via task-conditioning, we peek into HyperPrompt-Global models by looking at the distribution of attention scores. We choose the GLUE task MRPC as an example. To avoid biasing on individual examples, we aggregate over 100 validation examples to compute the quantity of interest (see \ref{appendix: attmass-entropy-calculation} for details). First, we compute the attention mass on hyper-prompts for each encoder layer. Figure \ref{fig: attention_vis} (top) shows that the network has lower attention mass on hyper-prompts in the lower layers and gradually increases attention mass for higher layers. This phenomenon indicates that higher-levels of Transformer becomes more task-specialized while it is beneficial for the lower-levels to learn task-agnostic representation \cite{yosinski2014transferable} by casting lower attention mass on hyper-prompts. Furthermore, we calculate the entropy of the attention scores on the tokens. For HyperPrompt-Global, we remove the hyper-prompts from the calculation and re-normalize the attention scores on the tokens to make a fair comparison with the MTL baseline. Figure \ref{fig: attention_vis} (bottom) shows a shift of entropy distribution towards higher values for HyperPrompt-Global. This signifies that injecting hyper-prompts encourages a more diverse attention distribution, which seems to be beneficial to model generalization.  


\begin{figure}[b]
  \centering
 \subfigure[Decoder]{
    \label{subfig: impact of prompt length on decoder} 
    \includegraphics[width=1.4in]{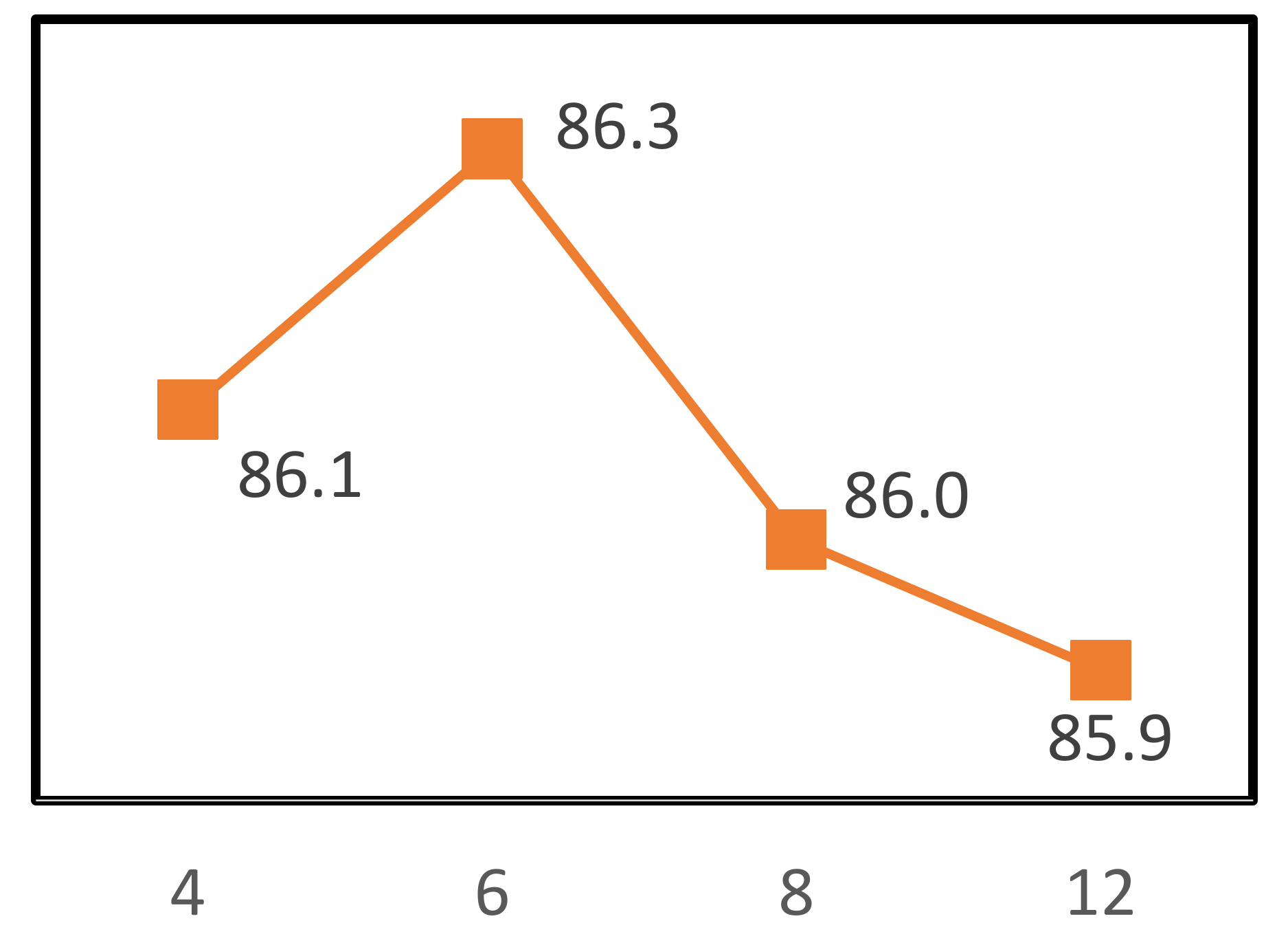}}
  \subfigure[Encoder]{
    \label{subfig: impact of prompt length on encoder} 
    \includegraphics[width=1.4in]{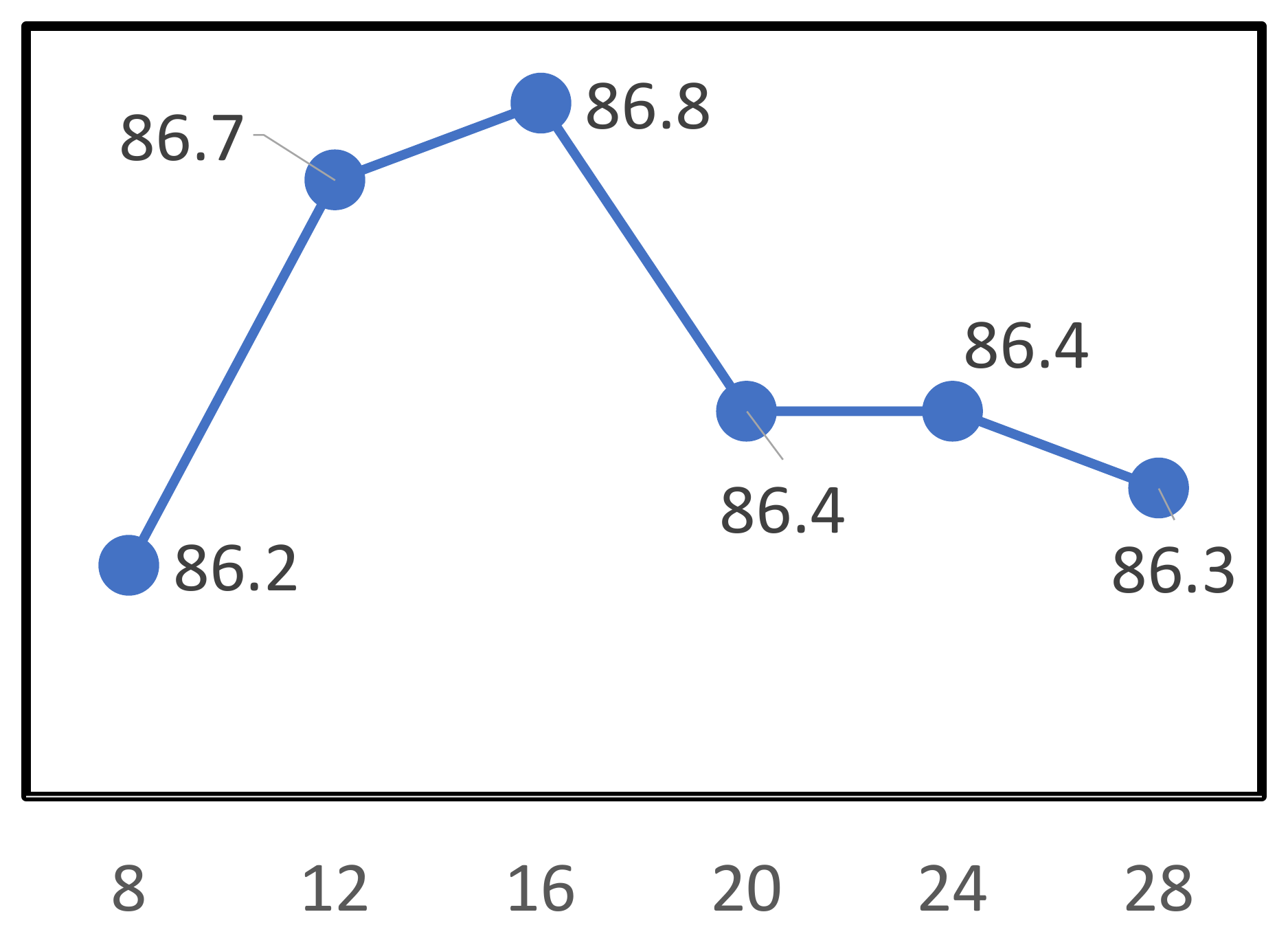}}
   \caption{Impact of hyper-prompt length in HyperPrompt-Global (GLUE score on T5 Base).}
  \label{fig: impact of prompt length} 
\end{figure}

\subsection{Impact of Hyper-Prompt Length}
\label{sec: Impact of Prompt Length}
%
%
%

HyperPrompt prepends $l$ trainable hyper-prompts to the keys and values of self-attention layer at every Transformer layer. In Figure \ref{fig: impact of prompt length}, we present the results of tuning the prompt length $l$ on GLUE using T5 Base as the example for HyperPrompt-Global (similar patterns are observed on T5 Large and SuperGLUE). We first add hyper-prompts on the decoder and search the best $l$ and then search the best $l$ for the encoder with the fixed best decoder hyper-prompt length. As shown in Figure \ref{subfig: impact of prompt length on decoder}, $l=6$ is the best for the decoder. As shown in Figure \ref{subfig: impact of prompt length on encoder}, HyperPrompt-Global achieves the best result of 86.8 when $l=16$ on the encoder with $l=6$ fixed for the decoder. The experiments show that hyper-prompts with length $l\sim \mathcal{O}(10)$ are good enough to achieve superior performance. Note that the original sequence length is 512 on the encoder and 32 on the decoder. Therefore, HyperPrompt does not substantially increase the time complexity of self-attention layers in practice.

\subsection{Encoder vs Decoder}

To understand the effect of adding task-conditioned parameters to different parts of the network, we present the results of HyperPrompt-Global and HyperFormer++ with adding hyper-prompts/adapters to: (1) encoder-only, (2) decoder-only, and (3) both encoder-decoder. As shown in Table \ref{tab:glue_hyperprompt_base_en_de}, we observe adding task-conditioned parameters to encoder (encoder-only) performs better than decoder-only on GLUE. However, the opposite is true for SuperGLUE, where encoder-only is substantially worse than decoder-only. This potentially could be a trainability issue when prompts are inserted into encoders, i.e. a different learning rate might be required to learn the prompt parameters from scratch.
We leave this investigation as a future work. Based on this experiment, we add task-conditioned parameters to the decoder for SuperGLUE in our experiments.

\begin{table}[ht]
    \centering
    \setlength{\tabcolsep}{2.0pt}
    \resizebox{0.48\textwidth}{!}
    {%
    \begin{tabular}{l|ccc}
    \toprule
    Model & \#Params  & GLUE & SuperGLUE \\
    \midrule
       MTL   &  1.0x  & 85.5 & 77.2\\
       HyperFormer++-Encoder & 1.02x  & 85.9 & 74.4\\
       HyperFormer++-Decoder & 1.02x  & 85.7 & 78.2\\
       HyperFormer++-Enc-Dec  & 1.04x & 86.5 & 74.8\\
       HyperPrompt-Encoder & 1.02x & 86.6 & 76.5\\
       HyperPrompt-Decoder & 1.02x & 86.3 & \textbf{78.9}\\
       HyperPrompt-Enc-Dec & 1.04x & \textbf{86.8} & 78.7\\
        
        \bottomrule
    \end{tabular}
    }%
    \caption{Ablation of inserting hyper-prompts or adapters into Encoder/Decoder/Enc-Dec (Base model).}
    \label{tab:glue_hyperprompt_base_en_de}
\end{table}

\section{Related Work}

\textbf{Prompt-Tuning.} Prompt tuning is becoming a new paradigm for adapting pre-trained general-purpose language models to downstream tasks, as a lightweight alternative to the popular fine-tuning approach. Here, we use the term Prompt-Tuning to cover a general family of methods following the prompting idea in GPT-3 \cite{brown2020gpt3}. To avoid manually design the prompts, recent efforts have focused on search for discrete prommpting words automatically \cite{shin2020autoprompt}. On the other hand, soft prompts \cite{li2021prefix,hambardzumyan2021warp,lester2021power,liu2021gpt} in the form of continuous vectors are introduced to simplify the process and have shown competitive results in both natural language understanding \cite{lester2021power, liu2021gpt} and generation tasks \cite{li2021prefix}. In particular, \citet{lester2021power} show that soft prompts can become competitive against full fine-tuning for a 11B parameters model, but with a big performance gap when the model size is moderate. In our work, we close this gap in the full fine-tuning setting and demonstrated that HyperPrompt can outperform strong multi-task baselines across all model sizes studied.

\textbf{Adapter-Tuning.} Adapter tuning \cite{houlsby2019adapter, houlsby2019parameter, karimi-mahabadi-etal-2021-parameter} is an alternative approach for parameter-efficient lightweight tuning of pre-trained langauge models for downstream tasks. Task-specific adapter layers \cite{houlsby2019adapter} are inserted into the Transformer block for fine-tuning while the rest of the backbone model is frozen. By adding only a few percent of additional parameters, \citet{karimi-mahabadi-etal-2021-parameter} show that competitive performance can be obtained on NLU benchmarks such as GLUE \cite{wang2018glue}. However, one limitation from the existing work is the evaluation of NLU on GLUE dataset, which is known to be no longer suitable for measuring the progress of language understanding \cite{wang2019superglue}. In our work, we evaluate HyperPrompt on SuperGLUE in addition to GLUE dataset, and show that indeed higher-difficulty tasks such as SuperGLUE requires full-tuning of the model beyond adapter tuning, to be competitive against state-of-the-art multi-task baselines. We also demonstrate that it is advantageous to inject prompts into self-attention than adding adapters.

\textbf{Multi-task Natural Language Understanding.} Multi-task learning is an important and challenge research direction in both full fine-tuning and prompt-tuning paradigms because of the competing needs of training and serving a single model while achieving Pareto efficiency in all tasks. 

The T5 model \cite{raffel2019exploring} renders all NLP tasks as a Text-to-Text problem. However, the best
results are obtained by task-specific fine-tuning.  MTDNN (multi-task deep neural network) \cite{liu2019MDNN}
shares parameters between several NLP tasks, and achieves strong performance on the GLUE
benchmark.
\citet{aghajanyan2021muppet} use around 50 tasks to boost the multi-task learning performance. \citet{aribandi2021ext5} builds an extremely diverse set of 107 NLP tasks for extreme multi-task scaling and demonstrate superior performances on a wide range of benchmarks.
Recently, \citet{wei2021flan, sanh2021multitask}
also illustrated how a multi-task learning stage can greatly improve the zero-shot prompting performance of large language models. 

\section{Conclusion}
We propose a novel architecture for prompt-based task-conditioning of self-attention in Transformers. The hyper-prompts are generated by a HyperNetwork to enable flexible information sharing among tasks while remain efficient in parameters and computation. HyperPrompt allows the network to learn task-specific feature maps where the hyper-prompts serve as task global memories, encouraging a more diverse distribution of attention. Extensive experiments show that HyperPrompt  can  achieve  superior  performances over strong T5 multi-task learning baselines and parameter-efficient models including Prompt-Tuning and HyperFormer++ on GLUE and SuperGLUE benchmarks.



\nocite{langley00}

\bibliography{icml2022}
\bibliographystyle{icml2022}

\newpage
\appendix
\onecolumn
\section{Appendix}

This section covers the parameter count of HyperPrompt, the experimental details, the calculation of attention mass and entropy, and per-task performance of GLUE and SuperGLUE.

\subsection{Parameter Count of HyperPrompt (\S\ref{subsec: para-efficiency})}
\label{appendix: para-efficiency}
Since the encoder and the decoder of Transformers have approximately the same capacity, the calculation considers only the decoder-side for simplicity. First, we have global task prompts $\bm{P}_{\tau} \in \mathbb{R}^{l \times d}$ for the $\tau$-th task, which contains $dlT$ parameters for $T$ tasks. The global HyperNetworks contain four weight matrices $\bm{W}^{D_{k}} \in \mathbb{R}^{(d \times b) \times t}$, $\bm{W}^{D_{v}} \in \mathbb{R}^{(d \times b) \times t}$, $\bm{W}^{U_{k}} \in \mathbb{R}^{(b \times h \times d_{h}) \times t}$ and $\bm{W}^{U_{v}} \in \mathbb{R}^{(b \times h \times d_{h}) \times t}$, which result in $4(bdt)$ parameters (we let $d = h \times d_{h}$). To obtain layer-aware task embedding, HyperPrompt learns task embedding $k_{\tau} \in \mathbb{R}^{t^{\prime}}$ for the $\tau$ task and layer embedding $z_{m} \in \mathbb{R}^{t^{\prime}}$ for the $m$-th Transformer block, which in total results in $Tt^{\prime}+Mt^{\prime}$ parameters. Besides, a task projection network $h_{t}$ is applied to fuse the task embedding and the layer embedding into the final layer-aware task embedding $\bm{I}_{\tau}^{m} \in \mathbb{R}^{t}$. $h_{t}$ is a two-layer feed-forward networks and contains $(2t^{\prime} + t)e$ parameters, where $e$ is the hidden dimension for $h_{t}$. 

\subsection{Experimental Details (\S\ref{sec: Experimental Setup})}
\label{appendix: exp-details}
Our models were implemented using Mesh Tensorflow\footnote{\url{https://github.com/tensorflow/mesh}} \cite{shazeer2018mesh} with the T5 library\footnote{\url{https://github.com/google-research/text-to-text-transfer-Transformer}} \cite{raffel2019exploring}. Following \citet{raffel2019exploring}, all data are preprocessed as into a "sequence-to-sequence" format. The length of the sequence is 512 at the encoder and 32 at the decoder. For all experiments, we train models 300K steps with a batch size of 128 and each batch is a mixture which samples each task proportionately to the number of examples in the dataset. Learning rate is a constant of $1\text{e-}3$ with Adafactor optimizer \citep{shazeer2018adafactor}.


For hyper-parameters tuning, the length of prompt $l$ is selected from $\{12, 16, 20, 20, 24\}$ at the encoder and $\{2,4,6,8,10,12,14,16\}$ at the decoder. The bottleneck dimension $b$ in the transform matrices is set to $d/r$, where $d$ is the model dimension of the T5 models and $r$ is a reduction factor and selected from $\{16, 32, 64\}$. The dimension $t$ of the layer-aware task embedding is selected from $\{32, 64, 128\}$. 
For a fair comparison, the hyper-parameters of baseline methods are set to have approximately the same numbers of parameters as HyperPrompt with the exception that Prompt-Tuning and MTL-Prompt-Share are extremely parameter-efficient with significantly fewer parameters. 

\subsection{Attention Mass and Entropy calculation (\S\ref{sec:peeking})}
\label{appendix: attmass-entropy-calculation}
To calculate the attention mass over hyper-prompts per layer, we averaged the hyper-prompt attention softmax scores across 100 validation examples and each attention head in a layer, and summed across each query attending to the hyper-prompts. In other words, we aggregated the amount of attention given to hyper-prompts by queries.
To calculate the attention entropy over tokens (other than hyper-prompts), we calculated the entropy of the attention distributions (averaged across attention heads) for 100 validation examples. This results in $\sum_{n=1}^{100}\sum_{L=1}^{12}{|X_n|}$ entropies calculated and visualized in Figure \ref{fig: attention_vis} (bottom).
For the HyperPrompt model, this involved re-normalizing the softmax distribution after removing hyper-prompts, as we wanted to understand how the original tokens are attended to.

\subsection{Per-Task Performance of GLUE and SuperGLUE}
\label{appendix: per-task-perf}
Table \ref{tab:glue_tune_all_vs_specific_appendix} and \ref{tab:superglue_tune_all_vs_specific_appendix} below show the comparison of fine-tuning the entire model against task-specific parameters only on GLUE and SuperGLUE datasets. Table \ref{tab:glue_hyperprompt_base} and \ref{tab:superglue_hyperprompt_base} show the detailed results of full-tuning of HyperPrompt against baselines on T5 Base. Table \ref{tab:glue_hyperprompt_large} and \ref{tab:superglue_hyperprompt_large} show the detailed results of full-tuning of HyperPrompt against baselines on T5 Large.
\begin{table*}[ht]
    \centering
    \resizebox{\textwidth}{!}{%
    \begin{tabular}{l|ccccccccc|c}
    \toprule
    Tunable Parameters & Model & CoLA & SST-2 & MRPC & SST-B & QQP & MNLI & QNLI & RTE  & AVG \\
    \midrule
    All & MTL &  59.4 & 96.6 & 93.3/90.7 & 90.6/90.4 & 89.8/92.3 & 90.8/90.8 & 95.2 & 90.8 & 88.3\\
    All & HyperFormer++-T5.1.1$_{\textsc{large}}$ & 63.3	& 96.6	& 93.2/90.7 &	92.1/91.9 &	89.7/92.3 &	90.5/90.7 &	95.1 &	89.9 & 88.8\\
    All & HyperPrompt-T5.1.1$_{\textsc{large}}$ & 64.6	& 96.7	& 94.0/91.8 &	91.3/91.4 &	90.0/92.4 &	90.8/91.0 &	95.4 &	91.9 & 89.4\\
   Task-Specific & HyperFormer++-T5.1.1$_{\textsc{large}}$ & 58.9 &	95.7 &	92.7/90.0	& 91.6/91.5 &	87.7/90.7 &	89.8/90.0 &	94.5 &	87.0 & 87.3 \\
   Task-Specific & HyperPrompt-T5.1.1$_{\textsc{large}}$ & 57.5 &	96.7 &	93.6/91.2 &	91.9/92.0 &	87.0/90.1 &	90.3/90.6 &	95.0 &	87.7 & 87.5 \\
    \midrule
        
        \bottomrule
    \end{tabular}
    }%
    \caption{Comparison of fine-tuning all vs task-specific parameters on GLUE.}
    \label{tab:glue_tune_all_vs_specific_appendix}
\end{table*}

\begin{table*}[ht]
    \centering
    \resizebox{\textwidth}{!}{%
    \begin{tabular}{l|ccccccccc|c}
    \toprule
    Tunable Parameters & Model & BoolQ & CB & COPA & MultiRC & ReCoRD & RTE & WiC & WSC  & AVG \\
    \midrule
    All & MTL  &  88.5 & 95.8/98.2 & 87.0 & 85.5/56.3 & 89.2/88.6 & 91.7 & 74.0 & 89.4 & 85.9\\
    All & HyperFormer++-T5.1.1$_{\textsc{large}}$ & 88.9	& 98.7/98.2	& 86.7 &	85.4/56.7 &	89.4/88.8 &	92.1 &	74.5 &	90.7 & 86.4\\
    All & HyperPrompt-T5.1.1$_{\textsc{large}}$ & 88.7 &	99.1/98.8 &	91.0	&85.0/55.6 &	89.8/89.1 &	91.3&	74.2 &	92.0 & 87.0\\
   Task-Specific &  HyperFormer++-T5.1.1$_{\textsc{large}}$ & 85.2 &	90.9/94.6 &	76.7	& 81.5/48.8 &	87.2/86.4 &	87.7&	67.8 &	82.1 & 80.5 \\
   Task-Specific &  HyperPrompt-T5.1.1$_{\textsc{large}}$ & 85.2 &	95.2/95.5 &	75.5	& 82.9/52.9 &	89.1/88.3 &	85.7&	71.1 &	82.2 & 81.5 \\
    \midrule
        
        \bottomrule
    \end{tabular}
    }%
    \caption{Comparison of fine-tuning all vs task-specific parameters on SuperGLUE.}
    \label{tab:superglue_tune_all_vs_specific_appendix}
\end{table*}

\begin{table*}[ht]
    \centering
    \resizebox{\textwidth}{!}{%
    \begin{tabular}{l|ccccccccc|c}
    \toprule
    Model & \#Params & CoLA & SST-2 & MRPC & SST-B & QQP & MNLI & QNLI & RTE  & AVG \\
    \midrule
       MTL  & 1.0x &  49.8 & 94.6 & 92.5/89.8 & 90.7/90.5 & 89.2/91.9 & \textbf{88.8}/\textbf{88.5} & 93.3 & 85.0 & 85.5\\
       Vanilla Adapter & 1.06x & \textbf{60.0}& \textbf{95.4}	& 92.7/89.8 &	90.2/90.2 &	\textbf{89.3}/91.9	& 88.5/88.1 &	\textbf{93.5} & 84.4 & 86.7\\
       HyperFormer++ & 1.04x & 56.9& 94.8	& 92.9/90.1 &	\textbf{91.1/90.9} &	88.9/91.7	& 88.7/88.3 &	93.4 & 	85.6 & 86.5\\
       Prompt-Tuning & 1.0003x & 48.0 & 95.0	& 92.2/89.0&	90.3/90.2 &	89.0/91.7	& \textbf{88.8}/\textbf{88.5} &	93.2 & 	82.9 & 84.8\\
       \midrule
      MTL-Prompt-Share (ours) & 1.008x & 56.2 & 94.7	& 93.0/90.4 &	90.6/90.4 &	89.2/91.9	& 88.7/88.4 &	93.4 & 	85.2 & 86.4\\
       MTL-Prompt-Sep (ours) & 1.06x & 57.2 & 94.6	& \textbf{93.8/91.4} &	91.0/90.8 &	89.2/91.9	& 88.5/88.4 &	93.4 & 	86.6 & \textbf{86.8}\\
       HyperPrompt (ours) & 1.04x & 57.0& 95.2	& 93.4/90.9 &	90.4/90.2 &	89.2/\textbf{92.0}	& 88.7/\textbf{88.5} &	93.4 & 	\textbf{87.1} & \textbf{86.8}\\
        
        \bottomrule
    \end{tabular}
    }%
    \caption{Comparison of HyperPrompt with baselines on GLUE using T5 Base.}
    \label{tab:glue_hyperprompt_base}
\end{table*}

\begin{table*}[ht]
    \centering
    \resizebox{\textwidth}{!}{%
    \begin{tabular}{l|ccccccccc|c}
    \toprule
    Model & \#Params &BoolQ & CB & COPA & MultiRC & ReCoRD & RTE & WIC & WSC  & AVG \\
    \midrule
       MTL  &  1.0x & 82.6 & 93.4/93.5 & 65.7 & 76.7/39.7 & 80.9/80.2 & 85.6 & 70.5 & 81.4 & 77.2\\
       Vanilla Adapter & 1.03x & \textbf{83.5} & 93.4/94.6	& 65.3 &	77.6/\textbf{42.7} &	81.0/80.2	& \textbf{88.2} &	71.0 & 	76.9 & 77.5\\
       HyperFormer++ & 1.02x & \textbf{83.5} & 96.2/97.0	& 66.3 &	\textbf{77.8}/41.9 &	81.2/80.4	& 87.4 &	71.0 & 	80.1 & 78.2\\
       Prompt-Tuning & 1.0003x & 82.5 & 94.0/95.8	& 68.0 &	76.9/40.2 &	80.9/80.2	& 84.1 &	69.3 & 	80.8 & 77.3\\
       \midrule
      MTL-Prompt-Share (ours)& 1.004x & 83.1& 95.7/95.2	& 67.7 &	77.3/41.3 &	\textbf{81.9}/81.0	& 87.4 &	70.4 & 	80.8 & 78.2\\
       MTL-Prompt-Sep (ours)& 1.03x & 83.3& \textbf{97.8}/\textbf{97.0}	& 61.7 &	77.6/42.3 &	81.5/80.6	& 86.8 &	\textbf{71.4} & 	78.2 & 77.5\\
       HyperPrompt (ours)& 1.02x & 83.3& 96.6/96.4	& \textbf{69.7} &	77.5/41.0 &	\textbf{81.7}/\textbf{80.9}	& 86.8 &	70.5 & 	\textbf{83.7} & \textbf{78.9}\\
        
        \bottomrule
    \end{tabular}
    }%
    \caption{Comparison of HyperPrompt with baselines on SuperGLUE using T5 Base.}
    \label{tab:superglue_hyperprompt_base}
\end{table*}

\begin{table*}[ht]
    \centering
    \resizebox{\textwidth}{!}{%
    \begin{tabular}{l|ccccccccc|c}
    \toprule
    Model & \#Params & CoLA & SST-2 & MRPC & SST-B & QQP & MNLI & QNLI & RTE  & AVG \\
    \midrule
       MTL  & 1.0x &  59.4 & 96.6 & 93.3/90.7 & 90.6/90.4 & 89.8/92.3 & 90.8/90.8 & 95.2 & 90.8 & 88.3\\
       Vanilla Adapter & 1.06x & 63.8& 96.5	& 93.7/91.3 &	92.0/\textbf{91.9} &	\textbf{90.0}/\textbf{92.5}	& 90.6/90.5 &	94.9 & 	88.7 & 88.8\\
       HyperFormer++ & 1.02x &63.3& 96.6	& 93.2/90.7 &	\textbf{92.1}/\textbf{91.9} &	89.7/92.3	& 90.5/90.7 &	95.1 & 	89.9 & 88.8\\
       Prompt-Tuning & 1.0001x & 62.5& \textbf{96.7}	& 93.4/91.0 &	91.3/91.0 &	\textbf{90.0}/92.4	& \textbf{90.9}/91.0 &	\textbf{95.4} & 	89.9 & 88.8\\
      MTL-Prompt-Share (ours) & 1.008x & \textbf{65.0} & \textbf{96.7} & 93.8/91.6 &	91.1/90.8 &	\textbf{90.0}/92.4	& 90.8/\textbf{91.1} &	95.3 & 	91.3 & 89.3\\
       MTL-Prompt-Sep (ours) & 1.06x & 63.9 & 96.6 & \textbf{94.6}/\textbf{92.6} &	92.0/91.7 &	\textbf{90.0}/92.4	& \textbf{90.9}/91.0 &	95.2 & 	91.6 & \textbf{89.4}\\
       HyperPrompt (ours) & 1.02x & 64.6& \textbf{96.7}	& 94.0/91.8 &	91.3/91.4 &	\textbf{90.0}/92.4	& 90.8/91.0 &	\textbf{95.4} & 	\textbf{91.9} & \textbf{89.4}\\
        
        \bottomrule
    \end{tabular}
    }%
    \caption{Comparison of HyperPrompt with baselines on GLUE using T5 Large.}
    \label{tab:glue_hyperprompt_large}
\end{table*}

\begin{table*}[ht]
    \centering
    \resizebox{\textwidth}{!}{%
    \begin{tabular}{l|ccccccccc|c}
    \toprule
    Model & \#Params & BoolQ & CB & COPA & MultiRC & ReCoRD & RTE & WIC & WSC  & AVG \\
    \midrule
       MTL  &  1.0x &88.5 & 95.8/98.2 & 87.0 & \textbf{85.5}/56.3 & 89.2/88.6 & 91.7 & 74.0 & 89.4 & 85.9\\
       Vanilla Adapter & 1.03x & 88.8& 98.3/\textbf{98.8}	& 86.0 &	85.3/56.0 &	89.3/88.7	& 91.2 &	73.6 & 	91.3 & 86.1\\
       HyperFormer++ & 1.01x &  \textbf{88.9} & 98.7/98.2	& 86.7 &	85.4/\textbf{56.7} &	89.4/88.8	& \textbf{92.1} &	\textbf{74.5} & 	90.7 & 86.4\\
       Prompt-Tuning & 1.0001x & 88.5 & 97.6/\textbf{98.8}	& 85.0 & 84.9/55.2 &	89.0/88.4	& 91.5 &	72.8 & 	90.1 & 85.6\\
       MTL-Prompt-Share (ours) & 1.004x & 88.5 & 98.7/98.2	& 88.0 &	85.2/55.8 &	89.7/89.1	& 91.8 &	74.1 & 	\textbf{93.9} & 86.8\\
       MTL-Prompt-Sep (ours) & 1.03x & 88.6 & 97.6/\textbf{98.8}	& 87.7 &	85.2/56.4 &	89.7/89.1	& 91.6 &	73.5 & 	89.4 & 86.1\\
       HyperPrompt (ours)& 1.01x  & 88.7& \textbf{99.1}/\textbf{98.8}	& \textbf{91.0} &	85.0/55.6 &	\textbf{89.8}/\textbf{89.1}	& 91.3 &	74.2 & 	92.0 & \textbf{87.0}\\
        \bottomrule
    \end{tabular}
    }%
    \caption{Comparison of HyperPrompt with baselines on SuperGLUE using T5 Base.}
    \label{tab:superglue_hyperprompt_large}
\end{table*}

\end{document}